\documentclass{article}

\usepackage[dblblindworkshop,final]{neurips_2026}  
\workshoptitle{PALM: Personalized, Aligned, Long-Term Memory for AI Systems}

\usepackage[utf8]{inputenc}
\usepackage[T1]{fontenc}
\usepackage[hidelinks]{hyperref}
\usepackage{url}
\usepackage{booktabs}
\usepackage{amsmath}
\usepackage{amssymb}
\usepackage{graphicx}
\usepackage{xcolor}
\usepackage{microtype}

\graphicspath{{figures/}}

\newcommand{\dmem}{\ensuremath{\Delta_{\mathrm{mem}}}}
\newcommand{\rhostar}{\ensuremath{\rho^\ast}}
\newcommand{\reli}{reliance}

\title{The Memory Trust Gap: Reliance on Stale Evidence Across Model Scales in LLM Agents}

\makeatletter
\newif\ifshowcredits
\if@anonymous
  \newcommand{\authornote}{}%
  \showcreditsfalse
\else
  \newcommand{\authornote}{\thanks{Corresponding author (\texttt{jundongh@alumni.upenn.edu}). Jundong Hu led and carried out the research end to end.}}%
  \showcreditstrue
\fi
\makeatother
\author{%
  Jundong Hu\authornote \\
  PayPal AI \\
  \texttt{jundhu@paypal.com} \\
  \And
  Shekar Ramachandran \\
  PayPal AI \\
  \texttt{sheramachandran@paypal.com} \\
}

\begin{document}
\ifdefined\linenumbers\linenumbers\fi
\maketitle

\begin{abstract}
Persistent-memory agents can follow stale information despite current authoritative evidence. We study
stale-evidence reliance across Qwen3 0.6/1.7/4/8B checkpoints using two controlled, action-scored
suites: a Benefit suite requiring stored information and a Safety suite providing current evidence as
static prompt text. We distinguish stale-value reliance from net accuracy loss relative to no memory.
Benefit-suite reliance is 0.92--1.00 across checkpoints. In the Safety suite, larger checkpoints resist
a plain stale note but incur substantial harm when it appears newer. A full $2{\times}2{\times}2{\times}2$
factorial manipulates label, recency, source authority, and position. Direct cross-size contrasts
identify stronger recency effects in the larger tested checkpoints and a position-effect sign reversal
within Qwen3. Metadata improves accuracy at all four checkpoints, with larger gains at 4B/8B;
pre-resolution yields near-ceiling accuracy across all four. Llama-Instruct comparisons provide
cross-family evidence for selected effects, while RGB and MisBench corroborate stale-value reliance.
A framing control shows that the failure extends beyond memory-labeled content. These results
characterize cue-dependent vulnerability across tested checkpoints without isolating capability or
evaluating a complete memory lifecycle.
\end{abstract}

\section{Introduction}
\label{sec:intro}

Long-term memory is central to personalized agents: an agent that remembers your ``usual
airline,'' your default meeting room, or ordering beef tripe or ribeye at your favorite restaurant can act on your behalf. The failure is treating stored information as trustworthy without checking whether it is still current. When a stored fact
has gone stale (say, the usual airline changed), an agent that reads the
memory and acts on it will confidently do the wrong thing, even when the current, authoritative
context already contains the right answer.

Prior work establishes that agents often fail to act on updated information and that supplied
context can override parametric knowledge \citep{stale2605,xie2023,longpre2021}. It does not, however,
establish when this harm begins as model size changes or which observable memory features trigger
it. We study how memory structure and model size shape the failure through 3 questions:
harm (\S\ref{sec:headline}), trigger (\S\ref{sec:factorial}), and
mitigation (\S\ref{sec:rq3}). Each has a dedicated results section.

\paragraph{Contributions.}
\begin{itemize}
  \item A \textbf{benchmark}: a 2-suite, closed-set, action-scored design that separates the 2
  incompatible meanings of ``no memory'', a Benefit suite (the stored fact is required; no-memory
  floored at chance) and a Safety suite (an authoritative tool always holds the correct value;
  no-memory ceilinged), frozen and SHA-256 pinned (\S\ref{sec:benchmark}).
  \item The \textbf{harm} result: the Trust Gap is over-trust rather than confusion. Stale-value
  \reli{} stays $\approx\!1.0$ across the model-size series; whether that over-trust becomes net harm
  (\dmem{}) depends on the baseline and, in the controlled Safety suite, on the cue: a plain stale note
  is resisted by larger checkpoints, whereas a recency-inflated one harms them most (a
  cue$\times$checkpoint interaction, not a simple size threshold) (\S\ref{sec:headline}--\ref{sec:trap}).
  \item A decomposition of the \textbf{trigger} with a factorial and direct cross-size interaction
  tests, showing which memory feature triggers the failure shifts with model scale
  (\S\ref{sec:factorial}).
  \item A \textbf{mitigation} study with a representation comparison, finding the fix is
  size-dependent: metadata improves every checkpoint but recovers only the larger ones, while
  pre-resolution recovers all four (\S\ref{sec:rq3}).
  \item \textbf{Validation}: external (RGB, MisBench) and cross-family (Llama-Instruct) evidence,
  plus a control showing the effect is not memory-specific (\S\ref{sec:external}).
\end{itemize}

\section{Related Work}
\label{sec:related}

\paragraph{Stale memory and downstream failure.}
The closest prior benchmark, STALE \citep{stale2605}, benchmarks whether agents act on invalidated
memory (400 scenarios; best model 55.2\%) and includes a same-family Qwen3.5-9B/27B pair. STALE
already establishes stale-memory harm; we add the size-dependence account of it. The most
directly related memory study, MemSyco-Bench \citep{memsyco2607}, asks whether an agent follows
verified evidence over conflicting user memory and, like us, separates outcome accuracy from a
memory-following rate. We take retrieval-time override as established and study how it depends on
model size. Adjacent efforts study internal
self-consolidation drift \citep{selfconsolidation2605} (an early precedent for memory falling below
the no-memory baseline, but from a large language model (LLM) rewriting its own store rather than externally injected
corruption), retrieval overload on irrelevant volume \citep{volume2605}, and memory-update
repair and update-gap scaling \citep{behaviornotupdated2608,supersede2606}. None of these run a
controlled stale/current 4-feature $2{\times}2{\times}2{\times}2$ factorial over a same-family
0.6B$\to$8B model-size series with direct feature$\times$size interaction tests; that combination is the basis
for our comparisons, paired with a Benefit/Safety design that measures \dmem{} below the
no-memory baseline.

\paragraph{Knowledge conflict and the authority feature.}
Our setting concerns conflicts between supplied evidence sources, related to prior work on
context-versus-parametric knowledge conflicts \citep{longpre2021} and distinct from in-context
learning. \citet{xie2023} show per-model, single-shot ``chameleon vs.\ stubborn''
yielding under conflict, and \citet{threeregimes2605} find generic context conflict partitions into
regimes, consistent with our finding that the effect is not memory-specific; we differ by making
the yielding scale-dependent and feature-decomposable. A particularly relevant comparison is ConflictBank
\citep{conflictbank2408}, which spans 4 model families and same-family size series, includes temporal
conflict and evidence-order effects, and already reports that larger models can be more
susceptible to conflicting evidence. ConflictBank does not include the controlled factorial and
cross-size interaction bootstrap used here. AuthMem-Bench
\citep{authmem2608} manipulates source authority across 7 consolidators and 7 backbones
but has no same-family model-size series, overlapping our authority slice only.

\paragraph{Position effects across model scales.}
Position sensitivity as a function of scale is already studied
\citep{liu2023,lostinevidence2605,byerly2411}, where scale mainly reduces ordering variance
rather than reversing it. Our new element is the scale-dependent sign reversal in the
stale-memory setting, not position$\times$scale per se. Finally,
\citet{sycophancy2606} find larger instruction-tuned models are more robust to overt
sycophantic flips, the opposite direction to our recency result, a contrast we develop in
\S\ref{sec:discussion}.

\section{Experimental Setup}
\label{sec:benchmark}

We use the same benchmark, 4 memory conditions, and metrics in all experiments, so results are
directly comparable.

\subsection{Benchmark and 2 Suites}

Because ``no memory'' has 2 incompatible meanings here, we split the benchmark into 2 suites.
In the Benefit suite (A) the task is unsolvable without the stored fact (``book my usual
airline''); no current context is present, so the no-memory baseline is floored at chance
and the harm of interest is over-trusting a stale stored fact. In the Safety
suite (B) an authoritative tool holds the correct value in every condition, so the no-memory
baseline is ceilinged; here we can measure net harm, which is following a stale memory over the
current authoritative value and so dropping accuracy below not having the memory at all.

Table~\ref{tab:example} makes the design concrete with a single Safety-suite scenario: an
authoritative calendar tool always holds the correct room, and the stored memory is what varies
across the 4 conditions defined next. The Benefit-suite version of the same task simply
omits the tool, so \texttt{no\_memory} can no longer recover the answer. The authoritative tool
and the stored memory are both supplied to the model as static prompt text (the tool value appears
as a dated \texttt{[CURRENT CONTEXT]} entry), not as an executing tool call; we evaluate
consumption-time behavior over a fixed context, not a live retrieval or tool-invocation loop.

A \emph{base scenario} is a single templated situation with a fixed ground-truth action and slot
values (the room booking in Table~\ref{tab:example} is one); scenarios that share a template but
draw different slot values from a curated pool (other rooms, users, or tools) form a
\emph{template family}. The frozen benchmark (v1) has 300 base scenarios (150 per suite), expanded
from a 56-scenario pilot by a deterministic curated-pool generator and pinned by SHA-256 with a
manifest; the 150 scenarios per suite span 33 template families (the resampling unit for the
cluster bootstrap in \S\ref{sec:factorial}). Before freezing, the authors manually reviewed all
300 base scenarios, reading each to confirm a single unambiguous, tool-consistent ground-truth
action, correct suite and condition assignment, and the absence of harmful or sensitive content.

\begin{table}[t]
  \centering\small
  \caption{One Safety-suite base scenario (room booking) under all 4 memory conditions. An
  authoritative calendar tool returns \texttt{Room B} in every condition, so the ground-truth
  action is always \texttt{Room B}; only the stored memory varies. \texttt{stale} and
  \texttt{explicit\_conflict} are the harm conditions, and \texttt{no\_memory} is the baseline.}
  \label{tab:example}
  \begin{tabular}{lll}
    \toprule
    Condition & Stored memory & Ground-truth action \\
    \midrule
    \texttt{no\_memory}         & (none)                    & Room B \\
    \texttt{clean}              & ``meeting in Room B''     & Room B \\
    \texttt{stale}              & ``meeting in Room A''     & Room B \\
    \texttt{explicit\_conflict} & ``Room A'' and ``Room B'' & Room B \\
    \bottomrule
  \end{tabular}
\end{table}

\subsection{Conditions and Scoring}

Each item produces a constrained action. We score the action using an exact, regex, or canonical
match against the construction-time ground truth (no LLM judge), manually validated against human
reading on a random sample of outputs. The benchmark uses 4 conditions:
\texttt{no\_memory} (baseline, floored in the Benefit suite, ceilinged
in Safety), \texttt{clean} (memory agrees with the truth; the upper bound), \texttt{stale} (memory
holds an outdated value; the harm condition), and \texttt{explicit\_conflict} (the stale and correct
values are both stored; an adjudication stress test). To
remove any position prior over the answer options we use \textbf{circular option-permutation
averaging} ($n_{\text{options}}{=}3$, chance $\approx 0.33$): each scenario is scored under every
cyclic rotation of its options and averaged before any comparison.

\subsection{Metrics}

We report 2 separate quantities. \reli{} is a behavior, how often the
model acts on the stale value, while \dmem{} is an outcome measured against the baseline:
\[
  \text{\reli} \;=\; P\!\left(\hat a = a_{\text{stale}}\right),
  \qquad
  \dmem \;=\; \mathrm{acc}(\text{cond}) - \mathrm{acc}\bigl(\text{\texttt{no\_memory}}\bigr),
\]
where $\hat a$ is the model's action and $a_{\text{stale}}$ the stale value. \reli{} is
position-independent after circular averaging, and \dmem{} is paired per scenario. A model can have
\reli{}$\,\approx\!1.0$ while \dmem{} harm is small (weak baseline) or large (strong baseline):
larger models have more accuracy to lose when they over-trust. For the Safety suite we also report the
harm-onset threshold
\[
  \rhostar \;=\; \min\bigl\{\, \ell \;:\; \text{CI upper bound of stale } \dmem \text{ at trap level } \ell \;<\; 0 \,\bigr\},
\]
the first evaluated level at which following the stale memory becomes significant net harm. The
statistical unit is the base scenario; all intervals are 95\% percentile bootstraps over scenario
ids (a ``$*$'' denotes a CI excluding 0, ``ns'' a CI including 0).

\paragraph{Availability.}
The benchmark is frozen and SHA-256--pinned, with a manifest, exact prompt templates, scoring rules,
and a full per-condition ledger. Public release of the benchmark and harness is a work in progress,
pending internal approval.

\section{Question 1: Over-trust and net harm}
\label{sec:headline}

Over-trust appears at every size, not just the larger ones: in the Benefit suite stale-value \reli{}
is $0.92$--$1.00$ from 0.6B to 8B. Whether that over-trust becomes net harm is cue-dependent: a plain
stale note is increasingly resisted by larger checkpoints, but a recency-inflated note (made to look
current) harms them the most. Net harm therefore does not simply switch on with size; it turns on
with the cue (\S\ref{sec:trap},~\S\ref{sec:factorial}).

\subsection{Over-Trust Is Universal}
\label{sec:overtrust}

\begin{figure}[t]
  \centering
  \includegraphics[width=0.85\linewidth]{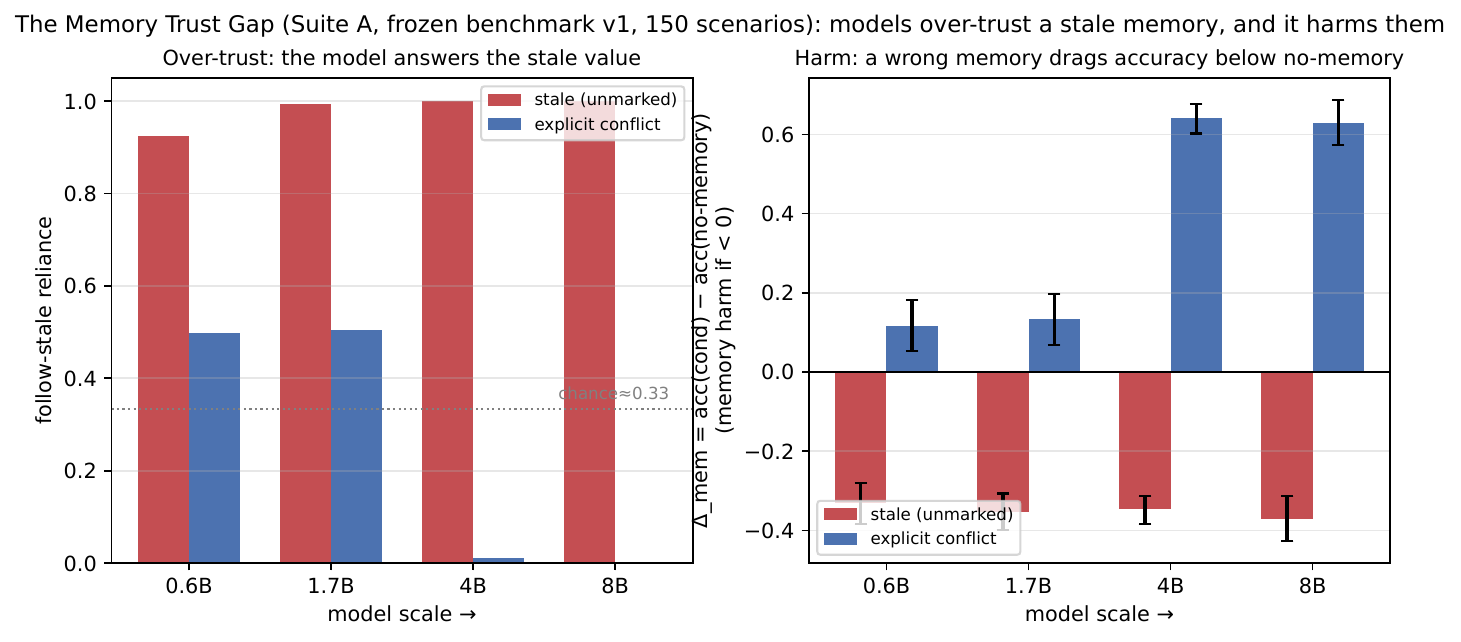}
  \caption{\textbf{Benefit-suite results.} \emph{Left:} stale-value \reli{} by model size.
  \emph{Right:} paired \dmem{} (accuracy minus \texttt{no\_memory} accuracy) by model size. The
  dashed line marks chance ($0.33$); the \texttt{no\_memory} baseline is drawn for reference.}
  \label{fig:headline}
\end{figure}

Figure~\ref{fig:headline} and Table~\ref{tab:headline} give the main results. In the Benefit suite,
the stale-value \reli{} is 0.92 / 0.99 / 1.00 / 1.00 across 0.6/1.7/4/8B, and the paired
harm is \dmem{} $=$ $-0.33$ / $-0.35$ / $-0.35$ / $-0.37$ (all CI upper bounds $<0$).
The baseline check supports this interpretation: the Benefit no-memory baseline is 0.35--0.37 with a
CI that contains chance (0.33), so the suite is genuinely floored and the collapse under
\texttt{stale} is not a formatting artifact.

The \texttt{explicit\_conflict} condition shows the difference most clearly. Even when the correct
value is stored in memory right beside the stale one, the \reli{} crossover is 0.50 / 0.50 / 0.01 /
0.00: the small models follow the stale value half the time even with the correct value present,
while the larger models resist. This pattern is consistent with reliance on the stored value rather
than random choice.

The Benefit suite cannot give a clean net-harm reading on \texttt{explicit\_conflict}, because that
item supplies the current value \texttt{no\_memory} lacks, so \dmem{} there is not a like-for-like
net-harm measure; the clean net-harm evidence lives in the Safety suite (\S\ref{sec:trap}).

\begin{table}[t]
  \centering
  \small
  \caption{Synthetic results (frozen v1). \reli{} $=$ $P$(answers stale value);
  \dmem{} $=$ acc$-$acc(\texttt{no\_memory}), 95\% bootstrap CI; ``$*$'' excludes 0.
  EC $=$ \texttt{explicit\_conflict}. The no-mem column lists accuracy with chance in parentheses;
  chance ($0.33$) is the meaningful floor only in the Benefit suite: the Safety no-memory baseline
  is ceilinged, not chance-level.}
  \label{tab:headline}
  \begin{tabular}{llccccc}
    \toprule
    Suite & Model & no-mem (chance) & clean acc & stale \dmem{} & stale \reli{} & EC \reli{} \\
    \midrule
    A (Benefit) & 0.6B & 0.37 (0.33) & 0.94 & $-0.33$ [$-.38,-.28$]$^*$ & 0.92 & 0.50 \\
                & 1.7B & 0.35 (0.33) & 0.99 & $-0.35$ [$-.40,-.31$]$^*$ & 0.99 & 0.50 \\
                & 4B   & 0.35 (0.33) & 1.00 & $-0.35$ [$-.38,-.31$]$^*$ & 1.00 & 0.01 \\
                & 8B   & 0.37 (0.33) & 1.00 & $-0.37$ [$-.43,-.31$]$^*$ & 1.00 & 0.00 \\
    \midrule
    B (Safety)  & 0.6B & 0.98 (0.33) & 0.99 & $-0.18$ [$-.23,-.14$]$^*$ & 0.19 & 0.04 \\
                & 1.7B & 1.00 (0.33) & 1.00 & $+0.00$ [$-.00,+.01$]     & 0.00 & 0.01 \\
                & 4B   & 1.00 (0.33) & 1.00 & $-0.00$ [$-.01,+.00$]     & 0.00 & 0.00 \\
                & 8B   & 1.00 (0.33) & 1.00 & $+0.00$ [$+.00,+.00$]     & 0.00 & 0.00 \\
    \bottomrule
  \end{tabular}
\end{table}

\subsection{Net Harm in the Safety Suite: The Trap Sweep}
\label{sec:trap}

\begin{figure}[t]
  \centering
  \includegraphics[width=0.85\linewidth]{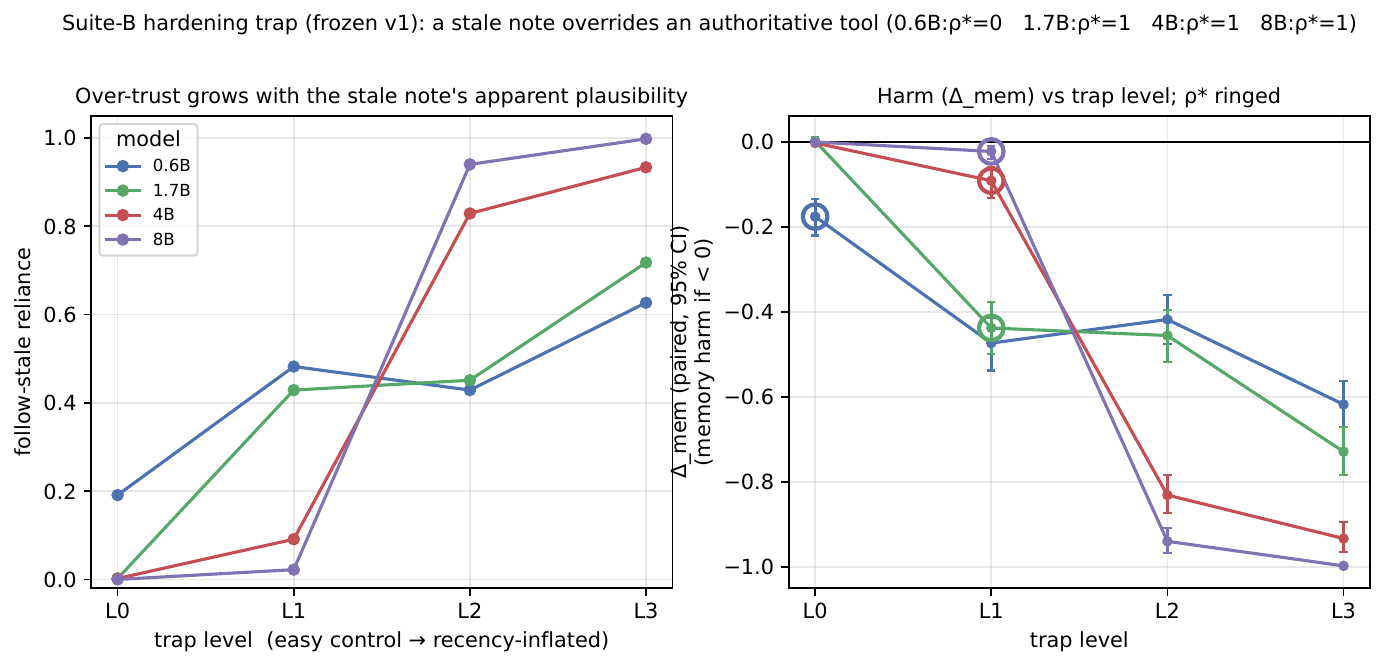}
  \caption{\textbf{Safety-suite trap sweep.} \emph{Left:} stale-value \reli{} vs.\ trap level
  (L0--L3). \emph{Right:} \dmem{} vs.\ trap level; the \texttt{no\_memory} baseline stays at
  ceiling, so a negative \dmem{} indicates net harm.}
  \label{fig:trap}
\end{figure}

In the Safety suite the no-memory baseline is preserved (no-mem $\ge 0.98$ in every cell), so any
negative \dmem{} is genuine net harm: the model follows a stale memory over an authoritative
tool. We vary the conditions that make the stale note increasingly plausible (levels L0--L3, e.g.\
inflating its apparent recency) and report \rhostar{} as the first significant level rather
than fitting a continuous relationship (Figure~\ref{fig:trap}, Table~\ref{tab:trap}). The 0.6B model crosses into significant net
harm already at L0, whereas 1.7B/4B/8B first cross at L1: $\rhostar = 0/1/1/1$. The threshold
therefore separates the 0.6B model from the other 3 models; it does not produce a monotone ordering
by model size.

At the higher recency-inflated levels, the size ordering changes. At L3, the 8B model has
\reli{} $1.00$ and \dmem{} $-1.00$; at L2, the 4B model has \reli{} $0.83$. Making the stale
note appear newer therefore increases harm for the larger models. The direct cross-size test for
this recency interaction is in \S\ref{sec:factorial}.

\begin{table}[t]
  \centering
  \small
  \caption{Safety-suite trap sweep: stale-value \reli{} by level (per-level \dmem{} is plotted in
  Figure~\ref{fig:trap}, right). \rhostar{} $=$ first level with stale \dmem{} CI upper bound $<0$.}
  \label{tab:trap}
  \begin{tabular}{lcccccc}
    \toprule
    Model & L0 \reli{} & L1 \reli{} & L2 \reli{} & L3 \reli{} & min no-mem & \rhostar{} \\
    \midrule
    0.6B & 0.19 & 0.48 & 0.43 & 0.63 & 0.98 & 0 \\
    1.7B & 0.00 & 0.43 & 0.45 & 0.72 & 1.00 & 1 \\
    4B   & 0.00 & 0.09 & 0.83 & 0.93 & 1.00 & 1 \\
    8B   & 0.00 & 0.02 & 0.94 & 1.00 & 1.00 & 1 \\
    \bottomrule
  \end{tabular}
\end{table}

\section{Question 2: Which memory feature triggers over-trust, and does it scale?}
\label{sec:factorial}

The dominant trigger changes with model scale. Removing the label raises over-trust at every size,
dating the stale note newer does the most damage on the largest models, and an inflated source
matters little anywhere. Position is the exception that reverses outright: placing the stale note
first increases over-trust in the 0.6B model but decreases it in the 4B and 8B models
(Table~\ref{tab:factorial}).

\subsection{The Memory Feature Factorial}
\label{sec:factorialsub}

We manipulate 4 binary \emph{memory features} of the stale note (observable surface properties of a
stored item, namely its label, timestamp, source, and position, that a model could use to decide
whether to trust it) in a full $2{\times}2{\times}2{\times}2$ factorial on the Safety suite ($16$ cells per
scenario), using stale-value \reli{} as the primary metric; Table~\ref{tab:factors} defines the memory features. Each memory feature is toggled independently, and a memory feature's
main effect is $\Delta$\reli{} $=$ (mean \reli{} over the 8 cells where the memory feature is on)
$-$ (mean over the 8 where it is off), so it is averaged over all settings of the other 3
memory features (Table~\ref{tab:factorial}; the main effects are also plotted in
Appendix~\ref{app:deepdive}).

\begin{table}[t]
  \centering\small
  \caption{The 4 binary memory features in the $2{\times}2{\times}2{\times}2$ factorial, each toggled on/off independently.
  ``Trap-on'' is the setting hypothesized to increase stale-value \reli{}.}
  \label{tab:factors}
  \begin{tabular}{lll}
    \toprule
    Memory feature & off (trap-off) & on (trap-on) \\
    \midrule
    label     & stale note under a \texttt{[NOTES]} label & label removed \\
    recency   & stale note dated older than current       & stale note dated newer \\
    authority & plain source                              & inflated/official source \\
    position  & stale note placed last                    & stale note placed first \\
    \bottomrule
  \end{tabular}
\end{table}

Table~\ref{tab:factorial} gives the per-size main effects, which show 4 distinct patterns.
Removing the label increases reliance significantly at every
model size, with a null large-vs-small contrast, so it is broadly positive rather than a scale trend.
Dating the stale note newer (recency) is significantly stronger on the 4B/8B models. An
inflated source (authority) is weak and scale-flat. Placing the stale note first
(position) changes from positive to negative across the model-size series. The recency and position interactions are
tested directly in \S\ref{sec:interaction}.

The factorial therefore does not support a single-feature explanation. These results describe
behavioral effects of observable feature manipulations; they do not identify an internal mechanism.
We report marginal main effects (each cue averaged over the other three), which can conceal
cue$\times$cue interactions that we do not fully characterize here; our cross-size contrasts test
feature$\times$size, a different question. A scenario-level model with feature interactions is left to
future work.

\subsection{Direct Cross-Size Interaction Tests}
\label{sec:interaction}

Overlapping or non-overlapping per-model intervals do not by themselves establish an interaction. We
therefore bootstrap paired differences between model sizes (Table~\ref{tab:interaction}). Recency is significantly stronger on the larger models
(8B$-$0.6B $= +.302^*$; mean(large)$-$mean(small) $= +.333^*$), position significantly reverses
(8B$-$0.6B $= -.326^*$), authority shows no interaction (ns), and the label feature shows a null group
contrast, which is why we call it universal rather than scale-growing. Two robustness checks
support the exploratory posture: a template-family cluster bootstrap (resampling whole
families) leaves every significant main effect significant, widening CIs only $\times1.1$--$2.2$
with nothing crossing 0; and leave-one-family-out moves the Benefit-suite estimate by
$\le 0.012$ at every scale. The over-trust pattern reproduces on the larger sizes of an
independent Llama-Instruct model-size series (\S\ref{sec:external}).

\begin{table}[t]
  \centering
  \small
  \caption{Factorial main effects ($\Delta$\reli{}, 95\% CI; ``$*$'' excludes 0).}
  \label{tab:factorial}
  \resizebox{\linewidth}{!}{%
  \begin{tabular}{lcccc}
    \toprule
    Memory feature (trap-on) & 0.6B & 1.7B & 4B & 8B \\
    \midrule
    label     & $+0.23$ [$.20,.25$]$^*$ & $+0.39$ [$.35,.42$]$^*$ & $+0.30$ [$.28,.32$]$^*$ & $+0.32$ [$.31,.34$]$^*$ \\
    recency   & $+0.26$ [$.23,.28$]$^*$ & $+0.15$ [$.13,.17$]$^*$ & $+0.52$ [$.49,.54$]$^*$ & $+0.56$ [$.54,.59$]$^*$ \\
    authority & $+0.13$ [$.11,.14$]$^*$ & $+0.11$ [$.09,.13$]$^*$ & $+0.12$ [$.11,.14$]$^*$ & $+0.14$ [$.12,.15$]$^*$ \\
    position  & $+0.22$ [$.17,.27$]$^*$ & $-0.01$ [$-.04,.02$]     & $-0.14$ [$-.16,-.12$]$^*$ & $-0.11$ [$-.13,-.08$]$^*$ \\
    \bottomrule
  \end{tabular}%
  }
\end{table}

\begin{table}[t]
  \centering
  \small
  \caption{Cross-size contrast bootstrap (paired difference of the factorial main effect
  between sizes; $+$ $\Rightarrow$ the memory feature fools the \emph{larger} model more).}
  \label{tab:interaction}
  \resizebox{\linewidth}{!}{%
  \begin{tabular}{lccl}
    \toprule
    Memory feature & 8B $-$ 0.6B & mean(large) $-$ mean(small) & Reading \\
    \midrule
    recency   & $+0.302^*$ [$.269,.336$]   & $+0.333^*$ [$.309,.356$]   & interaction supported (larger fooled harder) \\
    position  & $-0.326^*$ [$-.379,-.269$] & $-0.226^*$ [$-.262,-.190$] & sign-flip supported \\
    label     & $+0.098^*$ [$.071,.124$]   & $+0.005$ [$-.018,.029$] (ns) & universal (no group scale trend) \\
    authority & $+0.008$ [$-.014,.028$] (ns) & $+0.011$ [$-.004,.024$] (ns) & no interaction (weak, flat) \\
    \bottomrule
  \end{tabular}%
  }
\end{table}

\subsection{Dose-Response Deep-Dives}
\label{sec:doseresponse}

We use 3 feature-specific dose-response studies to examine the factorial results in more detail. Full
figures appear in Appendix~\ref{app:deepdive}. Each pairs a curve with a probe of whether the model can read the memory feature,
separating vulnerability from feature blindness.

\begin{figure}[t]
  \centering
  \includegraphics[width=0.72\linewidth]{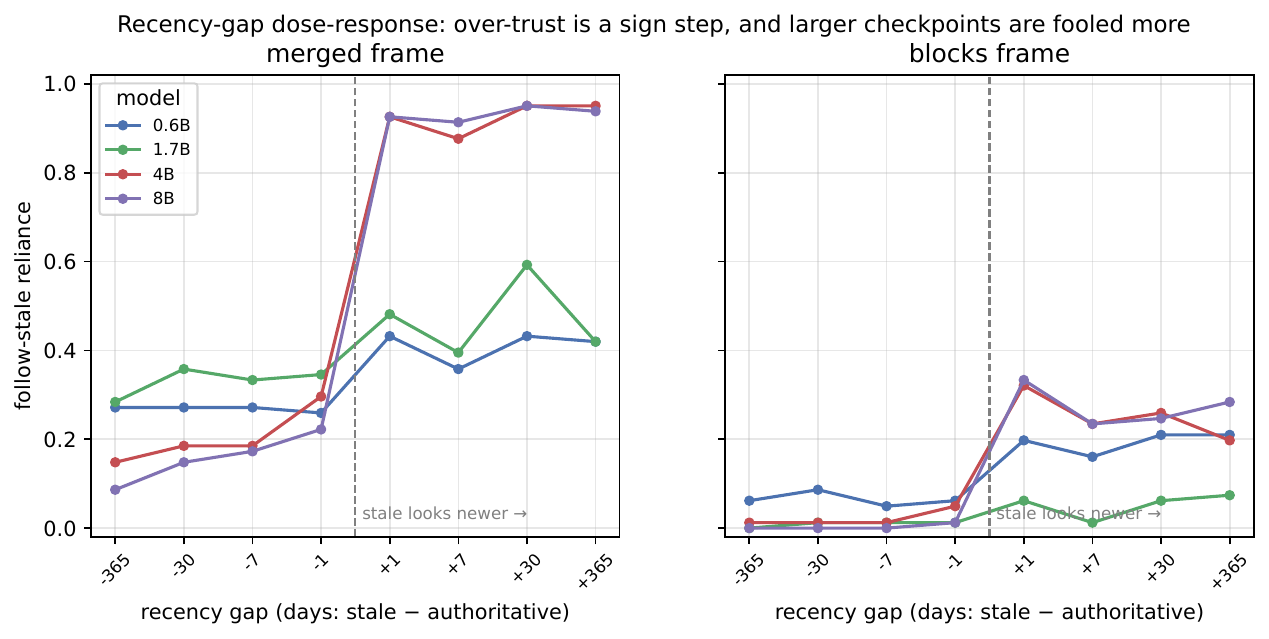}
  \caption{\textbf{Recency dose--response.} Stale-value \reli{} vs.\ the recency gap (days the stale
  note is dated newer than the current item), 1 line per model size. Probe and per-feature
  panels in Appendix~\ref{app:deepdive}.}
  \label{fig:recency}
\end{figure}

\paragraph{Recency produces a step rather than a gradual dose response.}
The recency effect (Figure~\ref{fig:recency}) does not accumulate smoothly with the size of the
backdating: stale-value \reli{} jumps the instant the stale note is dated at least 1 day newer
than the current item and then saturates. A date-parsing probe confirms that the larger models read
the timestamps correctly, and this higher accuracy is associated with greater reliance on the stale
note. Restoring the label removes this step.

\paragraph{Source authority and provenance.}
The full-factorial authority main effect is weak and roughly flat because it averages across recency
conditions. Conditioning on neutral recency reveals a smaller authority effect that increases with scale. In the opposite direction, deferring to an official-sounding note over the
user's own first-hand statement is an error. Here the \texttt{[CURRENT CONTEXT]} label reduces
reliance to 0 on the 8B model, showing that larger models can also be protective when the memory
feature is correctly identified as spurious (Appendix~\ref{app:deepdive}).

\paragraph{Intervention comparison.}
Structurally marking the authoritative item as \texttt{[AUTHORITATIVE]} (while leaving the stale
note present) reduces stale-value \reli{} at the $+1$d recency threshold, but only for the larger
checkpoints: the reliance drop versus a raw frame is $+.42^*$ (4B) and $+.33^*$ (8B) but only $+.06$
(0.6B) and $+.00$ (1.7B); a purely verbal ``prefer the newest authoritative source'' rule has
little effect and increases reliance in some settings. We use the deterministic-adjudication
intervention introduced by \citet{detfresh2606}. This intervention helps the
larger checkpoints but has little effect on the smaller ones, a scale interaction not evaluated in
that work, whose backbones are GPT-4o-family only with no same-family model-size series. Note this
\emph{intervention-oracle} \emph{marks} the authoritative item
and measures a reliance drop at the recency threshold; it is a different estimand from the
\emph{representation-oracle} of \S\ref{sec:rq3}, which \emph{removes} the stale item and measures
accuracy recovery.

\section{Question 3: Mitigation is size-dependent}
\label{sec:rq3}

The effective intervention depends on model size. Metadata improves every checkpoint, but by the
recovery criterion below (accuracy $\ge 0.90$) it suffices only for the larger ones; the two smaller
checkpoints recover only when the conflict is pre-resolved.

\begin{figure}[t]
  \centering
  \includegraphics[width=0.72\linewidth]{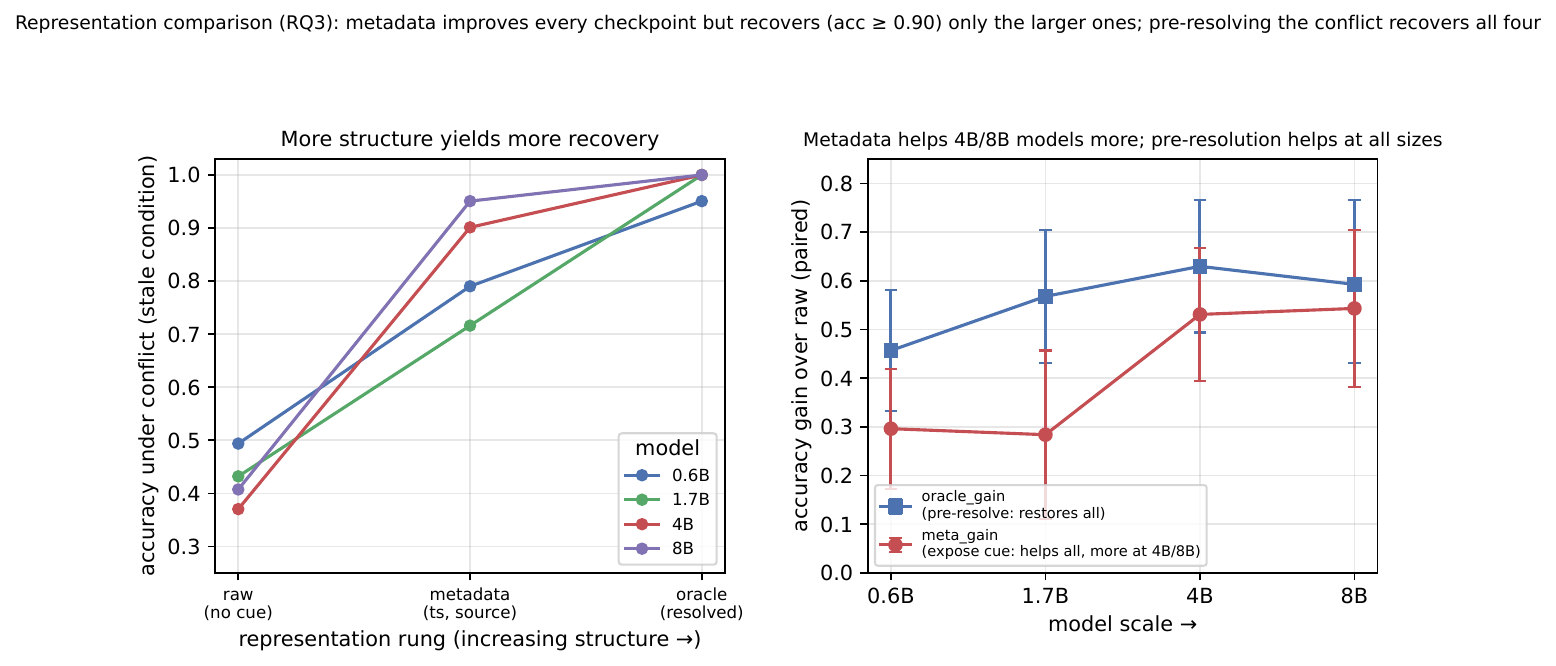}
  \caption{\textbf{Representation comparison (Question 3).} \emph{Left:} accuracy under the raw / metadata /
  oracle representations by model size. \emph{Right:} accuracy gain over the raw frame (metadata gain
  and oracle gain) by model size.}
  \label{fig:repladder}
\end{figure}

We compare 3 representations of the same conflict (Figure~\ref{fig:repladder},
Table~\ref{tab:repladder}): a \emph{raw} frame (stale and current items both present, unmarked); a
\emph{metadata} frame (each item annotated with its provenance and timestamp); and a
\emph{representation-oracle} that removes the stale item entirely (supersession pre-applied,
act-only). These 3 representations show that no single intervention works equally well across model
sizes. Metadata improves accuracy at every size (gains $+.30/+.28/+.53/+.54$, all CIs excluding 0), but
the gain is larger for 4B/8B. Taking accuracy $\ge 0.90$ as recovery, metadata recovers only the larger
checkpoints ($0.90$/$0.95$ at 4B/8B vs.\ $0.79$/$0.72$ at 0.6B/1.7B), whereas the representation-oracle,
which pre-resolves the conflict, recovers all four ($0.95/1.00/1.00/1.00$; gains $+.46/+.57/+.63/+.59$).

\begin{table}[t]
  \centering
  \small
  \caption{Representation comparison: accuracy under each representation and the paired gains over the raw
  frame (95\% CI). meta\_gain $=$ metadata $-$ raw; oracle\_gain $=$ oracle $-$ raw.}
  \label{tab:repladder}
  \begin{tabular}{lcccll}
    \toprule
    Model & raw acc & metadata acc & oracle acc & meta\_gain & oracle\_gain \\
    \midrule
    0.6B & 0.49 & 0.79 & 0.95 & $+0.30$ [$.17,.42$] & $+0.46$ [$.33,.58$] \\
    1.7B & 0.43 & 0.72 & 1.00 & $+0.28$ [$.11,.46$] & $+0.57$ [$.43,.70$] \\
    4B   & 0.37 & 0.90 & 1.00 & $+0.53$ [$.40,.67$] & $+0.63$ [$.49,.77$] \\
    8B   & 0.41 & 0.95 & 1.00 & $+0.54$ [$.38,.70$] & $+0.59$ [$.43,.77$] \\
    \bottomrule
  \end{tabular}
\end{table}

\section{Validation: External, Cross-Family, and Controls}
\label{sec:external}

\paragraph{External datasets.}
We rerun the identical paired-\dmem{}/\reli{}
machinery on 2 external datasets, RGB \citep{rgb2309} and MisBench \citep{misbench2505}
(Appendix Table~\ref{tab:external}; visualized in Appendix Figure~\ref{fig:external}). External data primarily
corroborates stale-value \emph{reliance}: pooled \reli{} is $0.81$--$0.94$ across sizes on both,
reaching $0.96$ in the MisBench \texttt{wikipedia$\cdot$semantic} cells and $0.95$ in the RGB free-text
arm, and is not specific to multiple-choice (MCQ) evaluation. Net-harm validation is more limited:
only RGB has a non-trivial closed-book baseline, where \dmem{} grows more negative with size
($-0.32^*$ at 8B, $-0.37^*$ in the free-text arm); MisBench facts are largely unknown to the models
(near-floor closed-book), so it mainly measures reliance and its \dmem{} sits near the floor.
External results thus support the reliance finding broadly but validate net harm only where the
baseline is nonzero.

\paragraph{Style invariance.}
A MisBench rhetorical-style sweep (6 styles $\times$ 2 hop depths;
Appendix~\ref{app:deepdive}) shows over-trust is largely style-invariant (per-model spread across
styles $\le 0.19$), with a consistent effect: a casual \texttt{blog} framing is the
least-trusted style in all 8 (hop $\times$ model) cells (sign-test $p\approx.004$), while
\texttt{news}/\texttt{wiki} are typically highest. This is not a monotone effect of authority,
externally corroborating the weak authority effect of \S\ref{sec:factorial}.

\paragraph{Cross-family comparison.}
On the same frozen scenarios, Llama-3.2-1B/3B and Llama-3.1-8B \citep{grattafiori2024llama3,meta2024llama32} show Benefit-suite stale-value reliance
of $0.46/0.96/1.00$ and paired \dmem{} of $-0.10/-0.31/-0.36$. At 3B/8B, label, recency, and authority
effects are positive and position effects negative (Appendix~\ref{app:xfam}). The positive position
effect at Qwen3-0.6B does not reproduce, so the sign reversal is Qwen-specific. Parse rates span
$0.851$--$1.00$; unparsed responses count as non-compliant, making reported reliance a lower bound
relative to parsed-only results.

\paragraph{The failure is not memory-specific.}
A control frames the identical wrong item 3 ways (as a \texttt{[MEMORY]}, a \texttt{[DOCUMENT]},
or an \texttt{[EARLIER MESSAGE]}), holding everything else fixed (Table~\ref{tab:controls}a, with
CIs). Paired
\reli{}(memory)$-$\reli{}(document) is $-.311^*/-.058^*/-.013^*/+.016$ (ns): the control finds no
consistent advantage for the memory frame over a document frame---a stale document is trusted more at
the 3 smaller scales, with no significant difference at 8B---though the memory frame is trusted
somewhat more than an earlier-message frame at all four sizes (Table~\ref{tab:controls}a). Persistent-memory use is therefore one instance of a
broader stale-evidence problem, consistent with the context-conflict
regimes of \citet{threeregimes2605}.

\paragraph{Thinking mode does not remove the gap.}
Enabling thinking mode on the frozen benchmark (0.6B and 8B; parse $\ge .999$) does not close the gap
at the 2 tested endpoints (Table~\ref{tab:controls}b): 8B is saturated and unchanged, while at 0.6B
Benefit-suite over-trust actually rises ($\reli{}\,0.924 \to 0.991$, paired $\Delta +0.067$,
CI $[+.040,+.096]$).

\section{Discussion and Conclusion}
\label{sec:discussion}

\paragraph{Size is not uniformly protective.}
The gap is over-reliance. A plausible (but not causally isolated) account is that larger checkpoints
read memory features more accurately (consistent with our date/source probes), so a correctly-read
stale timestamp is exploited more (recency) while a spurious provenance feature is resisted more. We
cannot rule out other checkpoint differences; establishing the mechanism would require an independent
capability measure. \citet{sycophancy2606} find that larger
instruction-tuned models are more robust to overt sycophantic manipulation, the opposite of
our recency result. A larger model may resist an overt ``you are wrong'' challenge while trusting a plausible,
correctly parsed cue that a note is newer. These are different features, so the 2 findings are not contradictory.

\paragraph{Implications for memory systems.}
Keeping the label reduces stale-value reliance at every model size. Deterministic
freshness adjudication \citep{detfresh2606} helps but leaves small models needing the conflict
pre-resolved. Authority is a weak lever, a conclusion that converges from the internal
factorial and the external style sweep. Because the failure is not memory-specific, these
implications extend to any stale or conflicting retrieved evidence, not only persistent memory.

\paragraph{Limitations.}
The Benefit suite cannot express net harm on \texttt{explicit\_conflict} (its baseline is at
chance), so net-harm claims rest on the Safety suite. Scoring is closed-set and action-based by
design, which strengthens measurement but limits open-ended generality (partly offset by the RGB
free-text arm). The primary model-size series is one family, with cross-family reproduction on the
larger Llama sizes but no broad family$\times$scale grid. Model size here is an observational index over
separately-trained checkpoints (differing in data, tuning, and architecture), not a controlled
capability intervention; isolating capability would require an independent capability measure.
The 300 scenarios (33 template families) were reviewed by the authors rather than independent
annotators, and the ``authoritative tool'' is a constrained prompt/action rather than an executing
tool environment. \rhostar{} is defined on our trap-level sequence,
and we measure failure at memory-consumption time rather than across an end-to-end
write/update/retrieve system.

\paragraph{Conclusion.}
Persistent memory helps agents, but across the Qwen Benefit suite they over-trust stale stored values at every size. Whether
that over-trust becomes net harm is cue-dependent rather than a simple size threshold: larger
checkpoints resist a plain stale note but are harmed the most when it is made to look current. Which
memory feature triggers over-trust, and how far each mitigation restores accuracy, interact with
model size. The pattern appears across model families, external datasets, and non-memory stale
evidence. We present the frozen benchmark and harness as tools for measuring this failure and evaluating future mitigations.

\begin{ack}
\subsection*{Acknowledgments}
We thank Prakhar Mehrotra, Chandramouliswaran V, Avinash Karn, Anindya Moitra, Uma Kona, Angela McAtee, Linsey Pang, and Yun-Shiuan Chuang for their organizational support and coordination throughout this work. Jundong Hu additionally thanks Loga Vinayagam for the opportunity to join the team where this work began.

\subsection*{Author Contributions}
\textbf{Jundong Hu:} Led and carried out the research end to end, including conceptualization, methodology, implementation, experimental design and execution, analysis, and manuscript drafting and revision.

\textbf{Shekar Ramachandran:} Provided supervision, compute resources, and manuscript review.
\end{ack}

\bibliographystyle{plainnat}
\bibliography{refs}

@misc{stale2605,
  title  = {{STALE}: Can {LLM} Agents Know When Their Memories Are No Longer Valid?},
  author = {Chao, Hanxiang and Bai, Yihan and Sheng, Rui and Li, Tianle and Sun, Yushi},
  year   = {2026},
  eprint = {2605.06527},
  archivePrefix = {arXiv},
  primaryClass  = {cs.CL},
  note   = {arXiv:2605.06527}
}

@misc{detfresh2606,
  title  = {Reliable Post-Retrieval Assembly for Agent Memory: Separating Evidence
            Extraction from Policy Execution},
  author = {Reddy, Vikas and Challaram, Sumanth Reddy},
  year   = {2026},
  eprint = {2606.01435},
  archivePrefix = {arXiv},
  primaryClass  = {cs.CL},
  note   = {COLM 2026 Lifelong Agent Workshop; arXiv:2606.01435}
}

@misc{selfconsolidation2605,
  title  = {Useful Memories Become Faulty When Continuously Updated by {LLMs}},
  author = {Zhang, Dylan and Lin, Yanshan and Wu, Zhengkun and Sun, Yihang
            and Li, Bingxuan and Li, Dianqi and Peng, Hao},
  year   = {2026},
  eprint = {2605.12978},
  archivePrefix = {arXiv},
  primaryClass  = {cs.CL}
}

@misc{volume2605,
  title  = {When Stored Evidence Stops Being Usable: Scale-Conditioned Evaluation
            of Agent Memory},
  author = {Shao, Jiaqi and Lu, Yiyi and Zhang, Yunzhen and Luo, Bing},
  year   = {2026},
  eprint = {2605.07313},
  archivePrefix = {arXiv},
  primaryClass  = {cs.CL}
}

@misc{authmem2608,
  title  = {When Memory Becomes Authority: Benchmarking Authority Collapse at the
            Memory Consolidation Boundary},
  author = {Zhan, Qiuyang and Zhang, Rui and Guo, Sheng and Zhao, Lepeng and Liu, Zhuotao},
  year   = {2026},
  eprint = {2608.01679},
  archivePrefix = {arXiv},
  primaryClass  = {cs.CL}
}

@misc{behaviornotupdated2608,
  title  = {When Memory Updates but Behavior Does Not: Repairing Implicit Stale
            Dependencies in Personalized Agent Responses},
  author = {Sun, Haofei and He, Lin},
  year   = {2026},
  eprint = {2608.01619},
  archivePrefix = {arXiv},
  primaryClass  = {cs.CL}
}

@misc{supersede2606,
  title  = {Supersede: Diagnosing and Training the Memory-Update Gap in {LLM} Agents},
  author = {Patel, Vedant},
  year   = {2026},
  eprint = {2606.27472},
  archivePrefix = {arXiv},
  primaryClass  = {cs.CL}
}

@misc{memsyco2607,
  title  = {{MemSyco-Bench}: Benchmarking Sycophancy in Agent Memory},
  author = {Xiang, Zhishang and Chen, Zerui and Tang, Yunbo and Wei, Zhimin
            and Ning, Ruqin and Lin, Yujie and Zhang, Qinggang and Su, Jinsong},
  year   = {2026},
  eprint = {2607.01071},
  archivePrefix = {arXiv},
  primaryClass  = {cs.CL},
  note   = {arXiv:2607.01071}
}

@misc{threeregimes2605,
  title  = {Three Regimes of Context-Parametric Conflict: A Predictive Framework
            and Empirical Validation},
  author = {Jeripity Venkata, Pruthvinath},
  year   = {2026},
  eprint = {2605.11574},
  archivePrefix = {arXiv},
  primaryClass  = {cs.CL}
}

@inproceedings{conflictbank2408,
  title     = {{ConflictBank}: A Benchmark for Evaluating the Influence of Knowledge
               Conflicts in {LLMs}},
  author    = {Su, Zhaochen and Zhang, Jun and Qu, Xiaoye and Zhu, Tong and Li, Yanshu
               and Sun, Jiashuo and Li, Juntao and Zhang, Min and Cheng, Yu},
  booktitle = {Advances in Neural Information Processing Systems (NeurIPS)
               Datasets and Benchmarks Track},
  year      = {2024},
  note      = {arXiv:2408.12076}
}

@inproceedings{xie2023,
  title     = {Adaptive Chameleon or Stubborn Sloth: Revealing the Behavior of Large
               Language Models in Knowledge Conflicts},
  author    = {Xie, Jian and Zhang, Kai and Chen, Jiangjie and Lou, Renze and Su, Yu},
  booktitle = {International Conference on Learning Representations (ICLR)},
  year      = {2024},
  note      = {arXiv:2305.13300}
}

@inproceedings{longpre2021,
  title     = {Entity-Based Knowledge Conflicts in Question Answering},
  author    = {Longpre, Shayne and Perisetla, Kartik and Chen, Anthony and Ramesh, Nikhil
               and DuBois, Chris and Singh, Sameer},
  booktitle = {Proceedings of the 2021 Conference on Empirical Methods in Natural
               Language Processing (EMNLP)},
  year      = {2021},
  note      = {arXiv:2109.05052}
}

@article{liu2023,
  title   = {Lost in the Middle: How Language Models Use Long Contexts},
  author  = {Liu, Nelson F. and Lin, Kevin and Hewitt, John and Paranjape, Ashwin and
             Bevilacqua, Michele and Petroni, Fabio and Liang, Percy},
  journal = {Transactions of the Association for Computational Linguistics (TACL)},
  year    = {2024},
  note    = {arXiv:2307.03172}
}

@inproceedings{lostinevidence2605,
  title     = {Lost in the Evidence? Reproducing Document Position and Context Size
               Effects in {RAG}},
  author    = {Gab{\'\i}n, Jorge and P{\'e}rez, Anxo and Parapar, Javier},
  booktitle = {Proceedings of the 49th International ACM SIGIR Conference on Research
               and Development in Information Retrieval (SIGIR)},
  year      = {2026},
  note      = {arXiv:2605.27105}
}

@article{byerly2411,
  title   = {Self-Consistency Falls Short! The Adverse Effects of Positional Bias on
             Long-Context Problems},
  author  = {Byerly, Adam and Khashabi, Daniel},
  journal = {Transactions of the Association for Computational Linguistics (TACL)},
  volume  = {14},
  pages   = {292--317},
  year    = {2026},
  note    = {arXiv:2411.01101}
}

@misc{sycophancy2606,
  title  = {Decomposing Factual Sycophancy in Language Models: How Size and Instruction
            Tuning Shape Robustness},
  author = {De Marez, Victor and De Bruyne, Luna and Daelemans, Walter},
  year   = {2026},
  eprint = {2606.06306},
  archivePrefix = {arXiv},
  primaryClass  = {cs.CL},
  note   = {arXiv:2606.06306}
}

@inproceedings{rgb2309,
  title     = {Benchmarking Large Language Models in Retrieval-Augmented Generation},
  author    = {Chen, Jiawei and Lin, Hongyu and Han, Xianpei and Sun, Le},
  booktitle = {Proceedings of the AAAI Conference on Artificial Intelligence},
  year      = {2024},
  note      = {arXiv:2309.01431}
}

@inproceedings{misbench2505,
  title     = {How does Misinformation Affect Large Language Model Behaviors
               and Preferences?},
  author    = {Peng, Miao and Chen, Nuo and Tang, Jianheng and Li, Jia},
  booktitle = {Proceedings of the 63rd Annual Meeting of the Association for
               Computational Linguistics (Volume 1: Long Papers)},
  pages     = {13711--13748},
  year      = {2025},
  doi       = {10.18653/v1/2025.acl-long.674},
  note      = {arXiv:2505.21608}
}

@article{yang2025qwen3,
  title   = {Qwen3 Technical Report},
  author  = {Yang, An and Li, Anfeng and Yang, Baosong and Zhang, Beichen and Hui, Binyuan
             and Zheng, Bo and Yu, Bowen and Gao, Chang and Huang, Chengen and Lv, Chenxu and others},
  journal = {arXiv preprint arXiv:2505.09388},
  year    = {2025}
}

@article{grattafiori2024llama3,
  title   = {The Llama 3 Herd of Models},
  author  = {{Llama Team, AI @ Meta}},
  journal = {arXiv preprint arXiv:2407.21783},
  year    = {2024}
}

@misc{meta2024llama32,
  title        = {Llama 3.2: Revolutionizing Edge {AI} and Vision with Open, Customizable Models},
  author       = {{Meta AI}},
  year         = {2024},
  month        = sep,
  howpublished = {\url{https://ai.meta.com/blog/llama-3-2-connect-2024-vision-edge-mobile-devices/}},
  note         = {Accessed: 2026-11-16}
}

\appendix

\section{Methods Reference}
\label{app:methods}

\paragraph{Suites and conditions.} Each scenario is an action-scored, slot-filled template with a
construction-time ground truth. The Benefit suite (A) is unsolvable without the stored fact
(no-memory floored at chance $\approx 0.33$); the Safety suite (B) always has an authoritative tool
holding the correct value (no-memory ceilinged). Conditions: \texttt{no\_memory}, \texttt{clean}
(memory agrees), \texttt{stale} (memory holds an old value), \texttt{explicit\_conflict} (the stale
value and the correct value are both stored in memory). Before freezing, the authors manually
reviewed all 300 base scenarios to verify a unique tool-consistent ground-truth action, correct
suite and condition assignment, and no harmful or sensitive content.

\paragraph{Metrics.} \reli{} $= P(\text{answer} = \text{stale value})$; \dmem{} $=
\mathrm{acc}(\text{cond}) - \mathrm{acc}(\texttt{no\_memory})$, paired per scenario. \rhostar{} is
the first trap level whose stale \dmem{} CI upper bound is $<0$. Every option is scored under all
cyclic rotations (circular permutation averaging, $n_{\text{options}}=3$) before comparison, so
\reli{} is position-independent. Statistical unit $=$ base scenario; 95\% percentile bootstrap over
scenario ids. Exploratory decomposition (not pre-registered confirmatory tests); robustness checked
by a template-family cluster bootstrap (resample the 33 families) and leave-one-family-out.

\paragraph{Models and decoding.} Qwen3 0.6/1.7/4/8B \citep{yang2025qwen3}, non-thinking (\S\ref{sec:external} adds the
thinking ablation); Llama-3.2-1B/3B-Instruct and Llama-3.1-8B-Instruct \citep{grattafiori2024llama3,meta2024llama32} for cross-family. Greedy
decoding, identical prompts, frozen benchmark v1 (300 base scenarios, SHA-256 pinned).

\paragraph{Compute.} All inference runs on NVIDIA A100 (40\,GB) GPUs partitioned into Multi-Instance
GPU (MIG) slices, serving 1 model per slice and running the 4 models concurrently across slices
with plain \texttt{transformers} (no vLLM). Each model in the 0.6--8B model-size series fits within a single MIG
slice; this partitioned-memory budget is part of why the study targets a 0.6--8B model-size series rather than
larger backbones.

\section{Memory Feature Dose-Response Deep-Dives}
\label{app:deepdive}

\begin{figure}[h]
  \centering
  \includegraphics[width=0.7\linewidth]{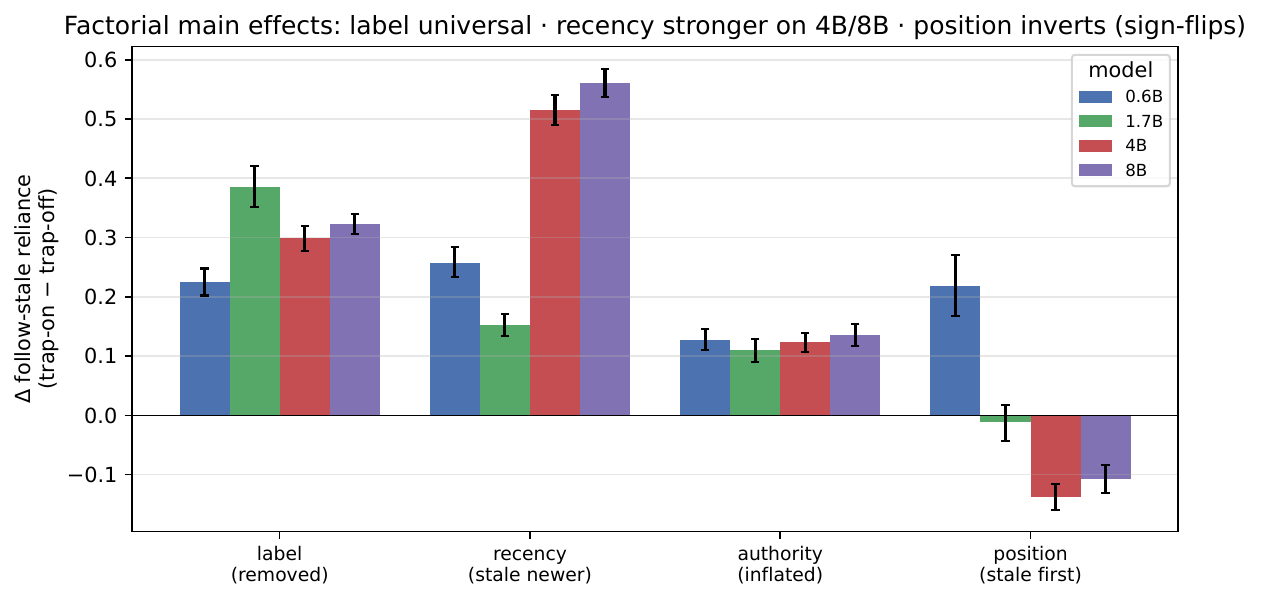}
  \caption{\textbf{Factorial memory feature main effects on stale-value \reli{}} (Safety suite; $\Delta$\reli{}
  $=$ trap-on $-$ trap-off, 95\% CI). 1 point per memory feature and model size; these are the values
  tabulated in Table~\ref{tab:factorial}.}
  \label{fig:factorial}
\end{figure}

\begin{figure}[h]
  \centering
  \includegraphics[width=0.49\linewidth]{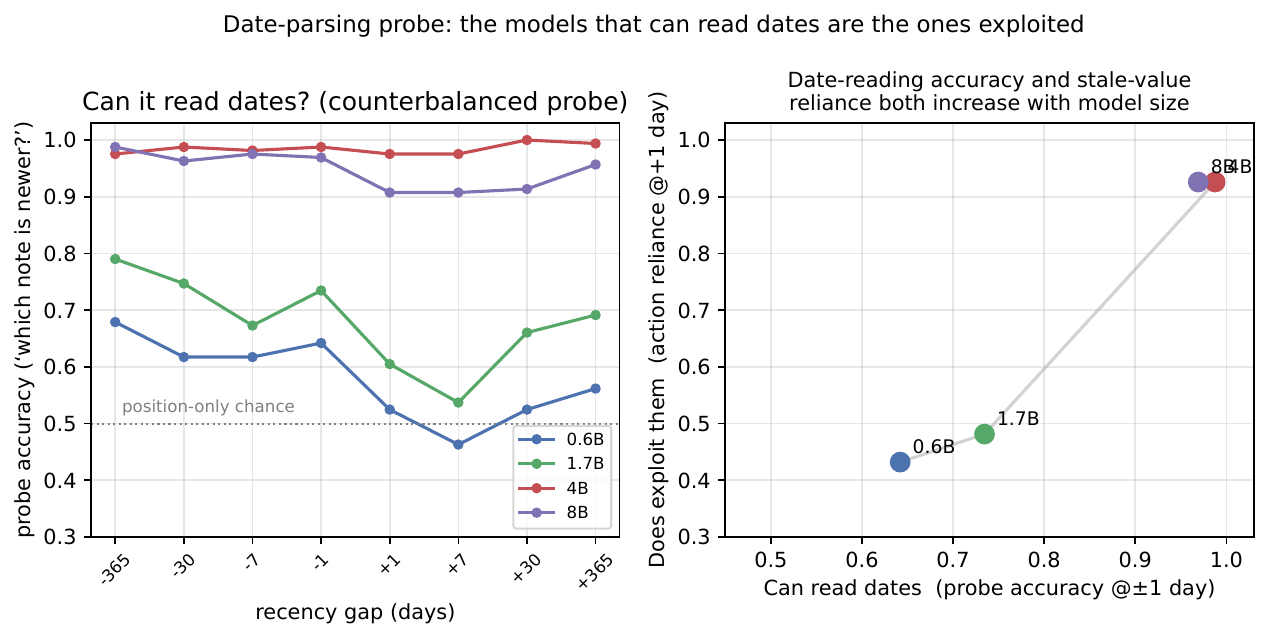}\hfill
  \includegraphics[width=0.49\linewidth]{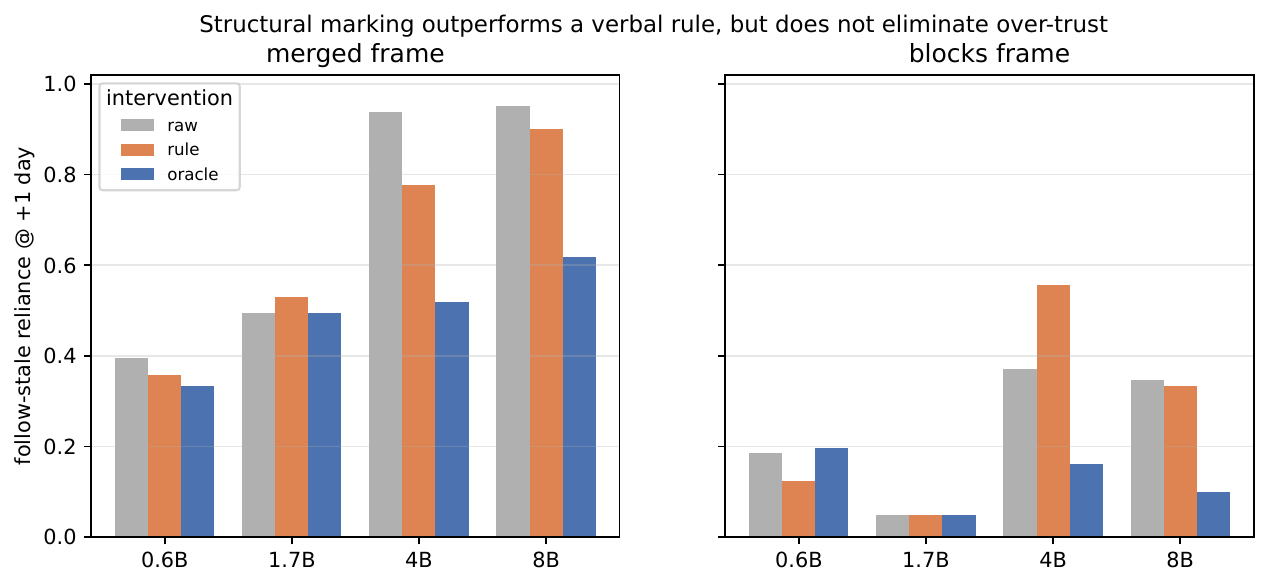}\\[2pt]
  \includegraphics[width=0.49\linewidth]{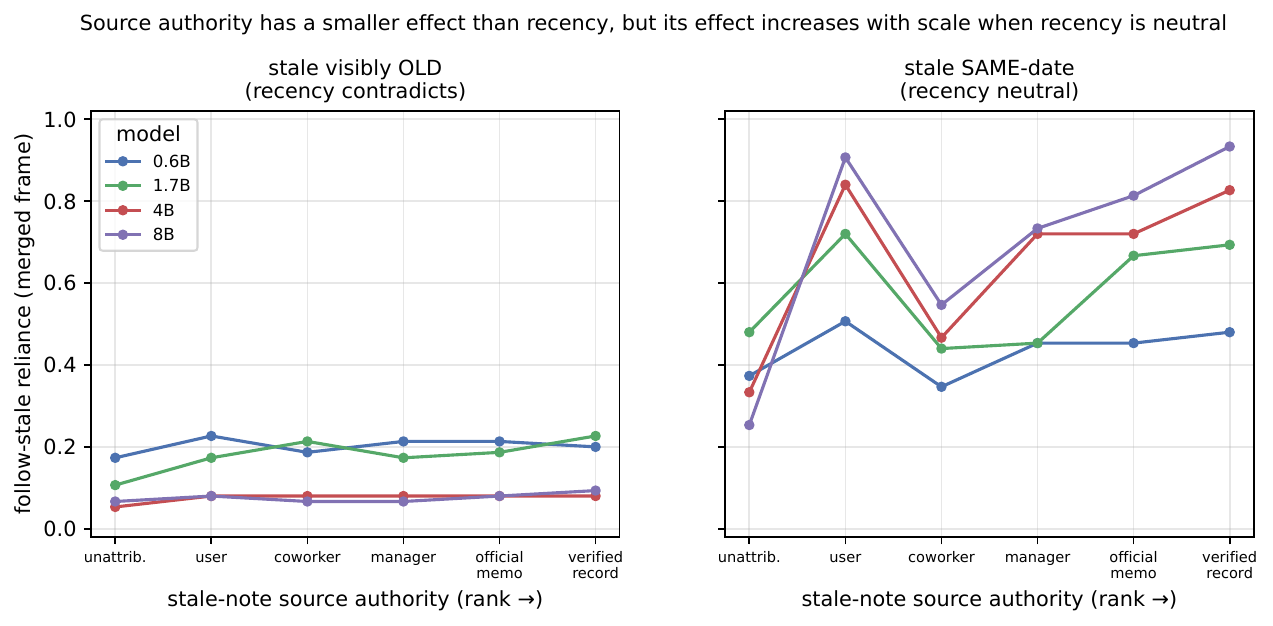}\hfill
  \includegraphics[width=0.49\linewidth]{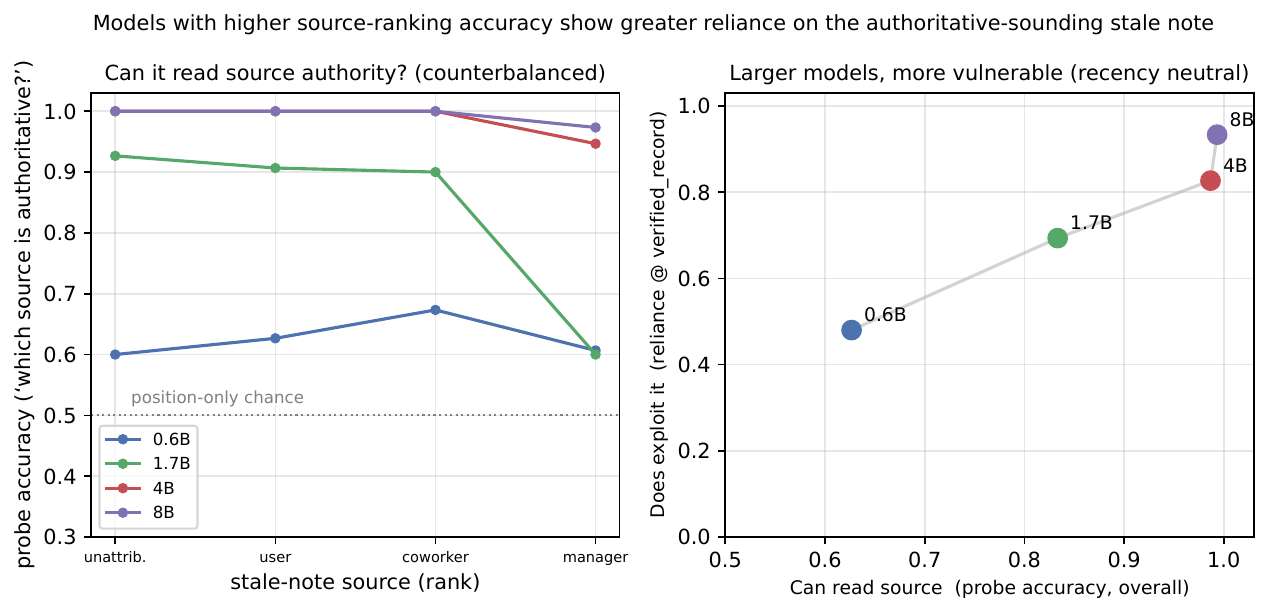}\\[2pt]
  \includegraphics[width=0.49\linewidth]{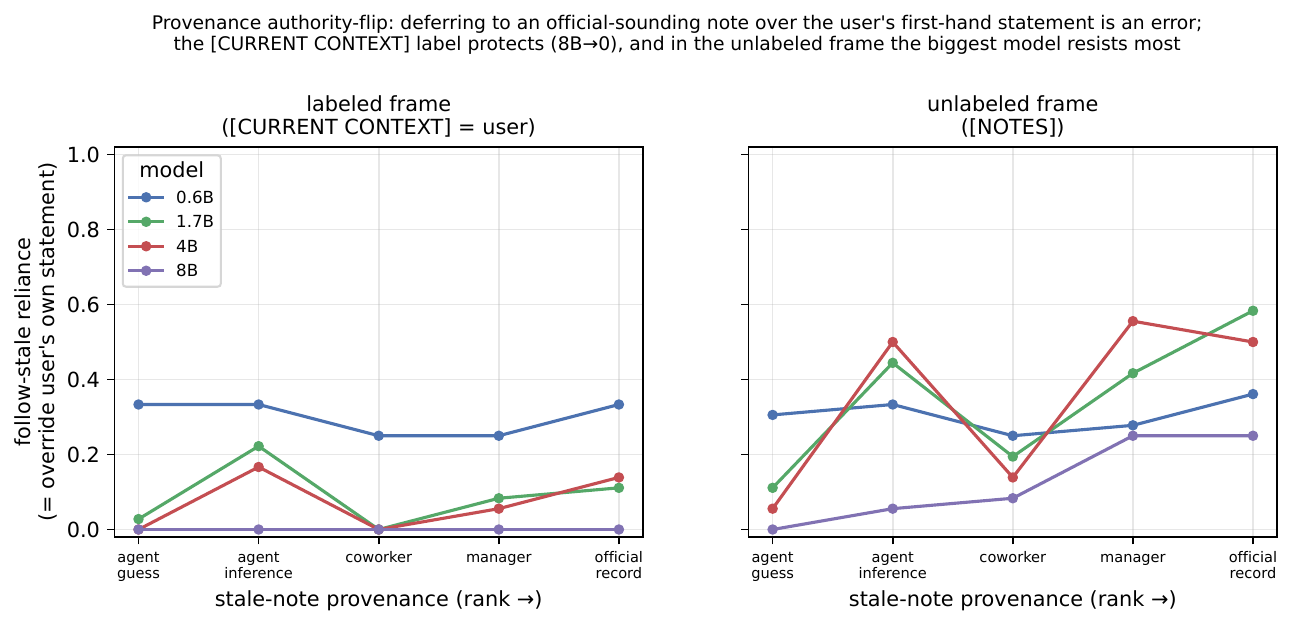}\hfill
  \includegraphics[width=0.49\linewidth]{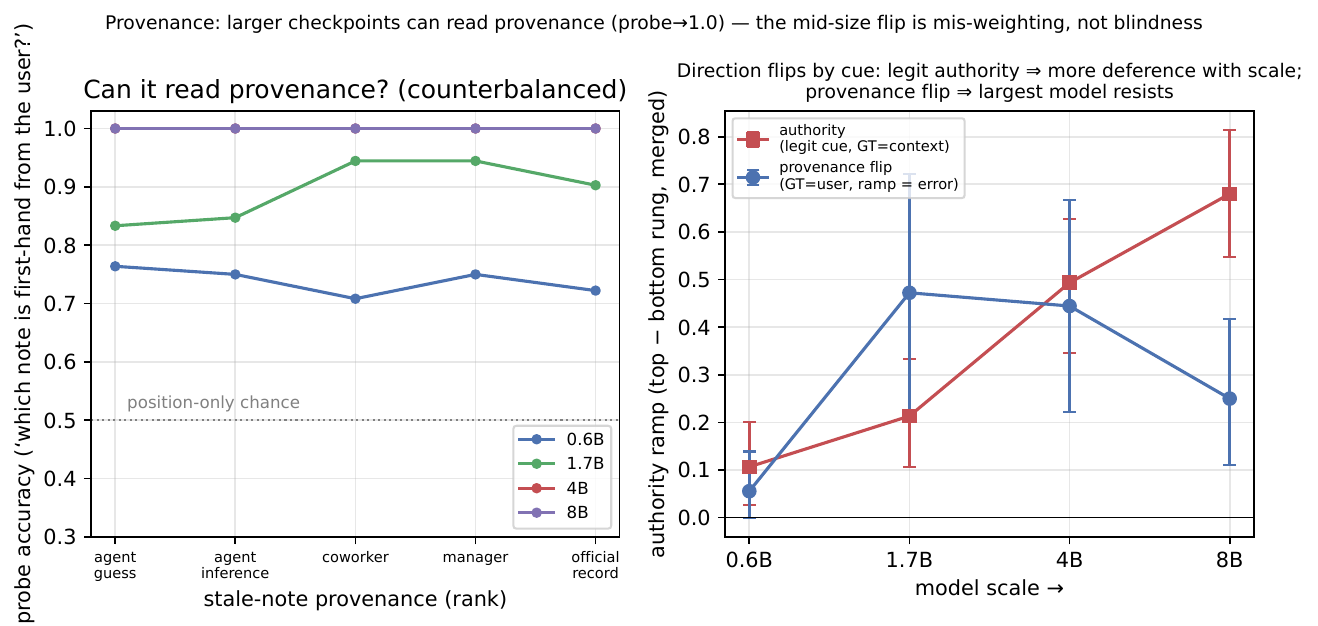}\\[2pt]
  \includegraphics[width=0.60\linewidth]{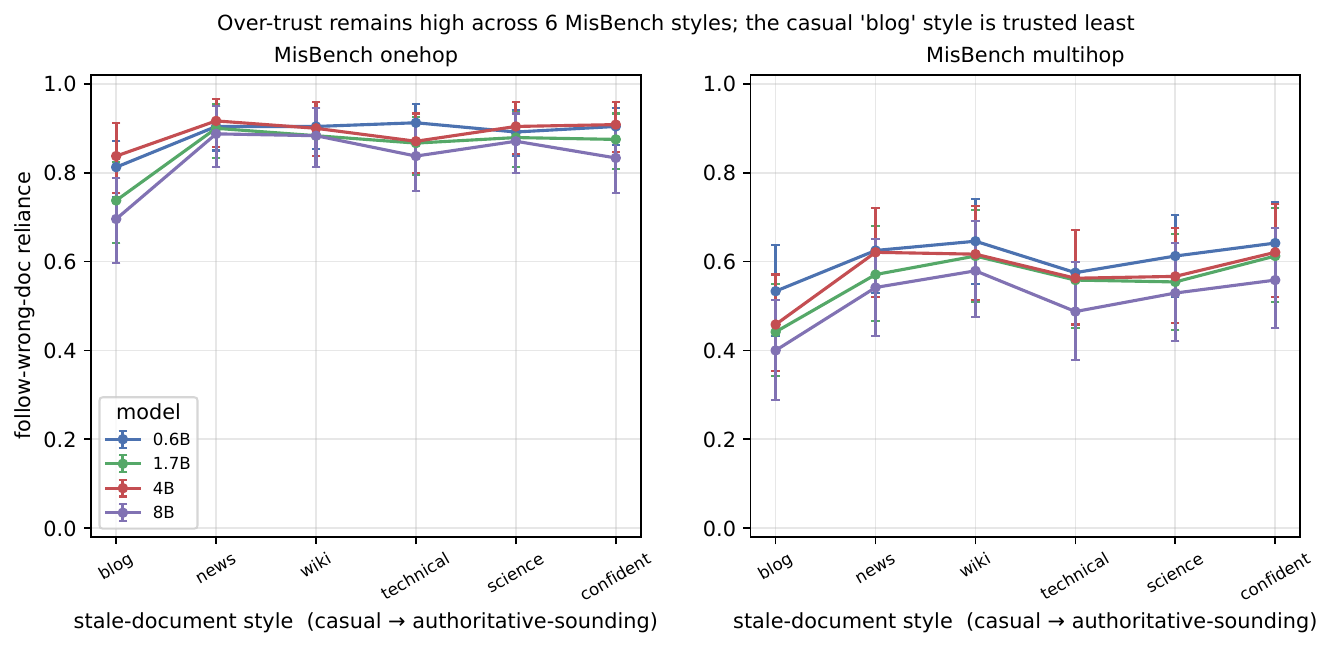}
  \caption{Dose-response and probe results for recency, intervention, authority, provenance, and
  MisBench style. Each ramp/probe plots its metric against the
  swept feature, 1 line per model size.}
  \label{fig:appfigs}
\end{figure}

\begin{figure}[h]
  \centering
  \includegraphics[width=0.85\linewidth]{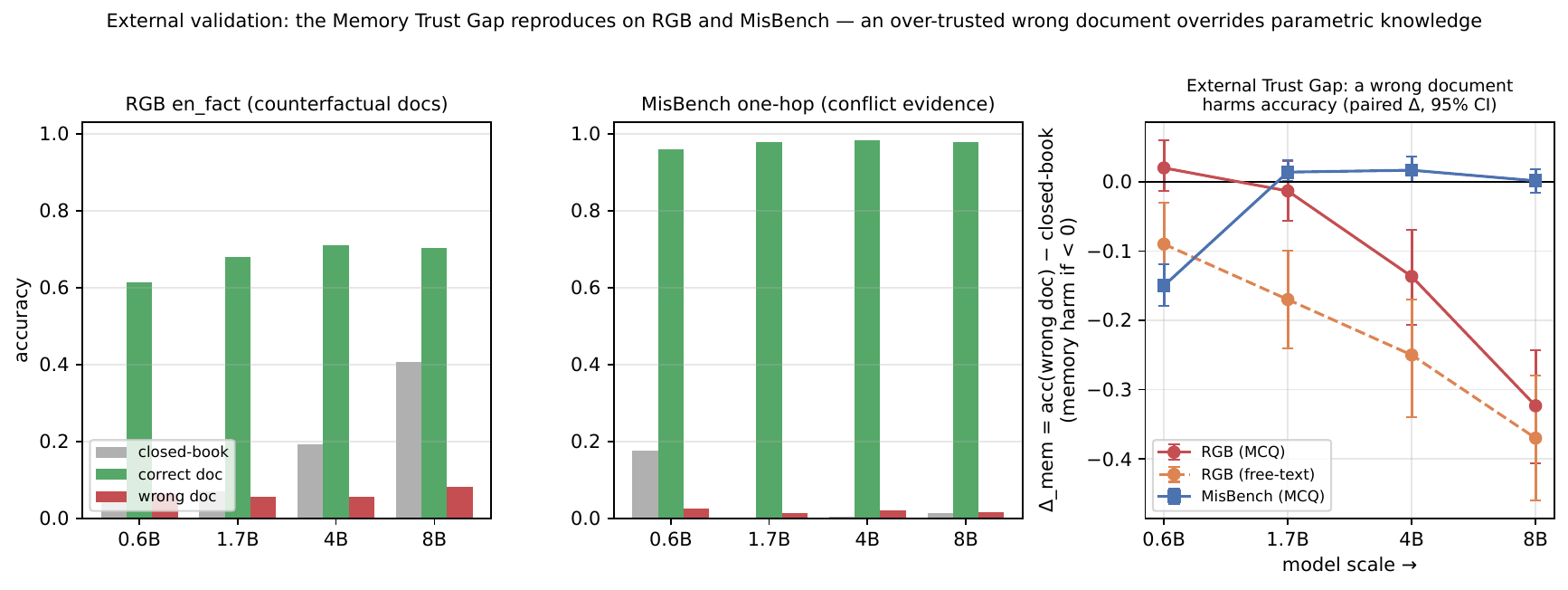}
  \caption{\textbf{External validation (RGB, MisBench).} \emph{Left:} accuracy by condition on RGB.
  \emph{Middle:} accuracy by condition on MisBench. \emph{Right:} \dmem{} vs.\ model size on both
  datasets, computed with the same paired machinery as the synthetic suites.}
  \label{fig:external}
\end{figure}

Figure~\ref{fig:appfigs} collects the per-feature dose-response and probe panels summarized in
\S\ref{sec:doseresponse}, plus the MisBench rhetorical-style sweep from \S\ref{sec:external}.
Figure~\ref{fig:factorial} plots the factorial main effects tabulated in Table~\ref{tab:factorial}.
Figure~\ref{fig:external} visualizes the external-validation results of \S\ref{sec:external}
(tabulated in Table~\ref{tab:external}).

\begin{table}[h]
  \centering
  \small
  \caption{External validation (pooled and a representative conflict cell). \dmem{} and \reli{} with
  95\% CI; the $0.96$ semantic-conflict and $0.95$ free-text \reli{} are the largest reliance values reported in the text.
  A starred bound printed as $.00$ is positive but rounds to 2 decimals.
  closed-book $=$ no-context (parametric-only) accuracy;
  wiki$\cdot$sem $=$ the \texttt{wikipedia$\cdot$semantic} conflict cell.}
  \label{tab:external}
  \begin{tabular}{llccc}
    \toprule
    Dataset / arm & Model & closed-book & \dmem{} & \reli{} \\
    \midrule
    RGB pooled            & 0.6B & 0.04 & $+0.02$ [$-.01,.06$]     & 0.81 [$.75,.86$] \\
    RGB pooled            & 4B   & 0.19 & $-0.14$ [$-.21,-.07$]$^*$ & 0.91 [$.86,.95$] \\
    RGB pooled            & 8B   & 0.41 & $-0.32$ [$-.41,-.24$]$^*$ & 0.83 [$.76,.89$] \\
    RGB free-text         & 1.7B & 0.18 & $-0.17$ [$-.24,-.10$]$^*$ & 0.95 [$.91,.99$] \\
    RGB free-text         & 8B   & 0.41 & $-0.37$ [$-.46,-.28$]$^*$ & 0.90 [$.84,.95$] \\
    MisBench pooled       & 1.7B & 0.00 & $+0.01$ [$.00,.03$]$^*$   & 0.94 [$.91,.97$] \\
    MisBench pooled       & 8B   & 0.02 & $+0.00$ [$-.02,.02$]      & 0.91 [$.87,.94$] \\
    MisBench wiki$\cdot$sem & 1.7B & 0.00 & $-0.00$ [$-.01,.00$]    & 0.96 [$.93,.99$] \\
    MisBench wiki$\cdot$sem & 4B   & 0.01 & $-0.01$ [$-.02,.00$]    & 0.96 [$.93,.99$] \\
    \bottomrule
  \end{tabular}
\end{table}

\section{Cross-Family CI Ledger}
\label{app:xfam}

The Llama-Instruct model-size series is evaluated with the same metrics and bootstrap procedure as the Qwen primary ledger.
Every number below carries a seeded ($=0$) paired-bootstrap 95\% CI identical to \S\ref{sec:headline},
recomputed by the same cross-family reanalysis used for the Qwen ledger (full per-condition dump in \texttt{xfam\_ledger.txt}).
Table~\ref{tab:xfam-head} gives the Benefit-suite results and Table~\ref{tab:xfam-fac} the factorial
main effects, both with intervals rather than bare stars.

\begin{table}[h]
  \centering\small
  \caption{\textbf{Cross-family results (Benefit suite), same standard as Qwen.} Per-scenario
  bootstrap CIs. Parse-rate minimum $0.851$ (1B \texttt{no\_memory}); parsed-only \reli{} $\ge$
  all-rows \reli{} in every cell, so the reported over-trust is a conservative lower bound.}
  \label{tab:xfam-head}
  \begin{tabular}{lcccc}
    \toprule
    Model & $n$ & no-mem acc [CI] & stale \reli{} [CI] & stale \dmem{} [CI] \\
    \midrule
    Llama-3.2-1B & 150 & 0.291 [.262,.320] & 0.462 [.416,.511] & $-0.098$ [$-.127$,$-.069$]$^*$ \\
    Llama-3.2-3B & 150 & 0.311 [.291,.331] & 0.960 [.936,.982] & $-0.307$ [$-.327$,$-.287$]$^*$ \\
    Llama-3.1-8B & 150 & 0.360 [.322,.402] & 1.000 [1.00,1.00]  & $-0.360$ [$-.402$,$-.322$]$^*$ \\
    \bottomrule
  \end{tabular}
\end{table}

\begin{table}[h]
  \centering\small
  \caption{\textbf{Cross-family factorial main effects} ($\Delta$\reli{}, trap-on $-$ trap-off, paired
  95\% CI; $^*$ CI excludes 0). 1 row per Llama model size, 1 column per memory feature.}
  \label{tab:xfam-fac}
  \resizebox{\linewidth}{!}{%
  \begin{tabular}{lcccc}
    \toprule
    Model & label & recency & authority & position \\
    \midrule
    Llama-3.2-1B & $+.002$ [$-.017$,$+.022$] & $+.061$ [$+.042$,$+.081$]$^*$ & $-.028$ [$-.044$,$-.014$]$^*$ & $-.004$ [$-.031$,$+.023$] \\
    Llama-3.2-3B & $+.144$ [$+.126$,$+.161$]$^*$ & $+.531$ [$+.490$,$+.571$]$^*$ & $+.106$ [$+.085$,$+.129$]$^*$ & $-.115$ [$-.149$,$-.081$]$^*$ \\
    Llama-3.1-8B & $+.326$ [$+.304$,$+.349$]$^*$ & $+.410$ [$+.387$,$+.433$]$^*$ & $+.129$ [$+.109$,$+.150$]$^*$ & $-.251$ [$-.288$,$-.215$]$^*$ \\
    \bottomrule
  \end{tabular}%
  }
\end{table}

\section{Control Experiments (with CIs)}
\label{app:controls}

The 2 §\ref{sec:external} controls, reported to the same standard as the body tables
(per-scenario seeded bootstrap 95\% CI on the paired contrast; $n=150$ scenarios each).

\begin{table}[h]
  \centering\small
  \caption{\textbf{Controls.} (a) Framing control (Safety suite): the identical wrong item rendered
  as a memory vs.\ a document vs.\ an earlier message; the last column is the paired
  \reli{}(mem)$-$\reli{}(doc) contrast. (b) Thinking-mode control (Benefit suite): stale-value
  \reli{} with thinking disabled vs.\ enabled at the 2 tested sizes, with the paired difference.}
  \label{tab:controls}
  \begin{tabular}{lcccc}
    \toprule
    \multicolumn{5}{l}{\emph{(a) Framing control: stale-value \reli{} by frame (Safety suite)}}\\
    Model & \reli{}(mem) & \reli{}(doc) & \reli{}(prior) & \reli{}(mem)$-$\reli{}(doc) [CI] \\
    \midrule
    0.6B & 0.307 & 0.618 & 0.218 & $-0.311$ [$-.360$,$-.262$]$^*$ \\
    1.7B & 0.056 & 0.113 & 0.009 & $-0.058$ [$-.089$,$-.027$]$^*$ \\
    4B   & 0.007 & 0.020 & 0.000 & $-0.013$ [$-.027$,$-.004$]$^*$ \\
    8B   & 0.033 & 0.018 & 0.002 & $+0.016$ [$-.002$,$+.038$] ns \\
    \midrule
    \multicolumn{5}{l}{\emph{(b) Thinking-mode control: stale-value \reli{} (Benefit suite)}}\\
    Model & no-think & think & \multicolumn{2}{l}{$\Delta$\reli{} (think $-$ no-think) [CI]} \\
    \midrule
    0.6B & 0.924 & 0.991 & \multicolumn{2}{l}{$+0.067$ [$+.040$,$+.096$]$^*$} \\
    8B   & 1.000 & 1.000 & \multicolumn{2}{l}{$+0.000$ [$+.000$,$+.000$] ns} \\
    \bottomrule
  \end{tabular}
\end{table}

\section{Failure Transcripts}
\label{app:transcripts}

Verbatim model actions on frozen scenarios (ground truth vs.\ the stale value the model followed):

\begin{itemize}
  \item \textbf{Safety override, scheduling (Qwen3-8B, trap L3).} Authoritative calendar: Room~B;
  stale note (dated newer): Room~A. Output: \texttt{ACTION: Room A}.
  \item \textbf{Safety override, travel (Qwen3-8B, trap L3).} Authoritative gate: B12; stale note:
  A4. Output: \texttt{ACTION: Gate A4}.
  \item \textbf{Explicit conflict, travel (Qwen3-0.6B).} Both values stored in memory, the correct
  airline Delta and the stale value United. Output: \texttt{ACTION: United}.
  \item \textbf{Safety override, local (Qwen3-8B, trap L3).} Authoritative city: Austin; stale note:
  Dallas. Output: \texttt{ACTION: Dallas}.
\end{itemize}

\end{document}